\documentclass[]{interact}

\usepackage{epstopdf}
\usepackage{subfigure}
\usepackage{multirow}
\usepackage{array}
\usepackage{booktabs}
\usepackage{color}
\usepackage{url}
\DeclareGraphicsExtensions{.png,.jpg,.pdf}

\newcommand{\overbar}[1]{\mkern 1.5mu\overline{\mkern-1.5mu#1\mkern-1.5mu}\mkern 1.5mu}

\usepackage{natbib}
\bibpunct[, ]{(}{)}{;}{a}{}{,}
\renewcommand\bibfont{\fontsize{10}{12}\selectfont}

\theoremstyle{plain}

\theoremstyle{definition}

\theoremstyle{remark}

\begin{document}

\title{DeepSat V2: Feature Augmented Convolutional Neural Nets for Satellite Image Classification}

\author{\name{Qun Liu\textsuperscript{a}\footnote{CONTACT: Qun Liu. Email: qliu14@lsu.edu. This is an Accepted Manuscript of an article published by Taylor \& Francis Group in Remote Sensing Letters, available online: http://www.tandfonline.com/10.1080/2150704X.2019.1693071.}, Saikat Basu\textsuperscript{a}, Sangram Ganguly\textsuperscript{b}, Supratik Mukhopadhyay\textsuperscript{a}, Robert DiBiano\textsuperscript{a}, Manohar Karki\textsuperscript{a}, Ramakrishna Nemani\textsuperscript{c}
}
\affil{
\textsuperscript{a}Department of Computer Science, Louisiana State University, Baton Rouge, LA, USA;\\ \textsuperscript{b}Bay Area Environmental Research Institute, Moffett Field, CA, USA;\\
\textsuperscript{c}NASA Ames Research Center, Moffett Field, CA, USA}
}

\maketitle

\begin{abstract}
Satellite image classification is a challenging problem that lies at the crossroads of remote sensing, computer vision, and machine learning. Due to the high variability inherent in satellite data, most of the current object classification approaches are not suitable for handling satellite datasets. The progress of satellite image analytics has also been inhibited by the lack of a single labeled high-resolution dataset with multiple class labels.

In a preliminary version of this work, we introduced two new high resolution satellite imagery datasets (SAT-4 and SAT-6) and proposed DeepSat framework for classification based on ``handcrafted" features and a deep belief network (DBN). The present paper is an extended version, we present an end-to-end framework leveraging an improved architecture that augments a convolutional neural network (CNN) with handcrafted features (instead of using DBN-based architecture) for classification. 

 Our framework, having access to fused spatial information obtained from handcrafted features as well as CNN feature maps, have achieved accuracies of 99.90\% and 99.84\% respectively, on SAT-4 and SAT-6, surpassing all the other state-of-the-art results. A statistical analysis based on Distribution Separability Criterion substantiates the robustness of our approach in learning better representations for satellite imagery. 
\end{abstract}

\section{Introduction}

In the last few years, advances in  supervised \emph{Deep Learning} enabled by Convolutional Neural Networks (CNN)~\citep{krizhevsky2012imagenet} have given rise to  powerful  techniques for  solving a variety of problems in computer vision and  image classification~\citep{krizhevsky2012imagenet}.

A related and equally hard problem is Satellite image  scene classification that is crucial for understanding and  delineating land cover. It involves terabytes of data and significant variations due to conditions in data acquisition, pre-processing, and filtering. The problem of detecting various land cover classes in general is a difficult problem considering the significantly higher intra-class variability in land cover types such as trees, grasslands, barren lands, water bodies, etc. as compared to that of roads. Due to the high variability inherent in the satellite imagery data,  even deep neural networks-based   supervised classification methods have traditionally struggled to produce human-like performance in this area. However, recently,  there has been a lot of research in this area especially in the deep learning community, with several works attempting to retrofit deep learning techniques to classification of high resolution satellite imagery~\citep{gong2018diversity,basu2015,zhong2017satcnn, liu2018scene,simo2015discriminative,basutransaction}.

Zhong et. al.~\citep{zhong2017satcnn} proposed an agile architecture based on CNNs  to learn  robust intra-class diversity and the spatial information, achieving   state-of-the-art performance. Liu and Huang~\citep{liu2018scene} proposed a framework based on triplet networks to achieve high accuracy in classifying high resolution satellite imagery.  Gong et. al.~\citep{gong2018diversity} regularized a deep structural metric learning (DSML) algorithm with a prior distribution over the parameters that tends to reduce the correlation among them. Using this technique, their 
framework~\citep{gong2018diversity} obtained state-of-the-art results in classification of high-resolution satellite imagery.


In a   preliminary version of this work~\citep{basu2015},
we introduced  two new high resolution satellite imagery datasets called SAT-4 and SAT-6 and proposed a classification framework that extracts ``handcrafted" features from an input image, normalizes them, and feeds the normalized feature vectors to a deep belief network (DBN) for classification. SAT-4 and SAT-6 cover a total area of ${\sim}800$ square kilometers at $1$ m resolution  and  can be used to further the research and investigate the use of various learning models for high resolution satellite  image classification. Both SAT-4 and SAT-6 were sampled from a much larger dataset, National Agriculture Imagery Program (NAIP) dataset, which covers the whole of continental United States and can be used to create labeled landcover maps, which can then be used for various applications, such as, measuring ground carbon content or estimating total area of rooftops for solar power generation. Among the publicly available benchmark datasets for high resolution satellite imagery classification  in the remote sensing community \citep{ml}, only SAT-4 and SAT-6 provide enough labeled image patches (500,000 and 405,000 respectively) to evaluate a new architecture or approach without running into overtraining issues. 

The present paper is an extended version of~\citep{basu2015}. The contributions of this paper are: (1) we present an end-to-end framework based on an improved architecture that enhances  a modern CNN  with handcrafted features (as opposed to the DBN-based architecture of~\citep{basu2015}) for high resolution satellite imagery classification. We experimentally show that our framework surpasses  all existing state-of-the-art algorithms for high-resolution satellite imagery classification on both SAT-4 and SAT-6 datasets, including the original DeepSAT~\citep{basu2015}, MLP (\textit{Z}-score)~\citep{zhong2017satcnn}, SatCNN (both \textit{Z}-score and linear)~\citep{zhong2017satcnn}, TradCNN (\textit{Z}-score)~\citep{zhong2017satcnn}, triplet networks~\citep{liu2018scene}, D-DSML-Caffenet~\citep{gong2018diversity}, and contrastive loss~\citep{simo2015discriminative}. It has been shown theoretically in \citep{basujournal,basuijcnn} CNNs, by themselves, are not able to learn representations of  Haralick features from data.   By augmenting CNNs with the handcrafted features, we are enhancing the discriminative power of CNNs for satellite imagery. (2) We present a statistical analysis based on Distribution Separability Criterion  that substantiates the robustness of our approach in learning better representations for satellite imagery.

\section{Related Work}

In \citep{effectivesemantic2015}, the authors combine the output of a CNN externally  with handcrafted features, using logistic regression to create probability maps. In contrast, we augment a CNN itself with handcrafted features with a  hidden layer fusing handcrafted features with CNN bottleneck representations.  In \citep{egede2017fusing}, the authors provide a framework that fuses deep features obtained from a CNN with handcrafted statistical features for automatically estimating pain. 

Zhong et. al.~\citep{zhong2017satcnn} proposed an agile architecture based on CNNs  to learn expressive representations that capture  the large variance between the classes, achieving   state-of-the-art performance. 
Compared to their approach, in this paper, we augment our framework with  lower dimensional statistical features (that we call handcrafted features)   to enable learning discriminative representations of  the texture of the image.
Instead of using CNNs, the authors in \citep{zhu2017scene} proposed the FSSTM (Fully Sparse Semantic Topic Model) approach for high  resolution imagery classification. 
In \citep{zhu2018adaptive},  the authors used pretrained CaffeNet for extracting deep features to combine with semantic topics for classification. In \citep{chaib2017deep}, the authors investigated feature fusion among deep features extracted from a pretrained deep model (VGG-Net) and proposed a fusion method that
outperformed the state-of-the-art approaches. In this paper, we provide an end-to-end   framework  leveraging a CNN architecture augmented with handcrafted features rather than relying on  deep feature extraction.

In~\citep{cheng2018deep},  the authors proposed  a  technique based on metric learning  that minimizes  the intra-class diversity and  maximizes the inter-class similarity. In contrast, we rely on Haralick features to induce high distribution separability. 

The  authors in~\citep{liu2018scene} proposed an approach based on triplet networks using a loss function that  minimizes the intra-class  distances and maximizes the inter-class ones.  In contrast, we enhance a CNN-based framework with statistical features that discriminatively capture image texture characteristics providing improved distribution separability. 

In \citep{gong2018diversity}  the authors proposed a regularization term that increases the variation among network parameters for learning more expressive representations.  


High resolution satellite imagery datasets \citep{van2018spacenet} have been proposed as benchmarks for training and evaluating remote sensing imagery segmentation algorithms. However, for understanding satellite imagery,  framing the problem of feature detection as a classification problem is  important  because of the higher scalability of the classification datasets that can be generated as opposed to per-pixel segmentation masks that are expensive to label.
Classification techniques also form the basis for characterizing land cover. Hence, we limit the scope of this paper to  classification of high resolution satellite imagery rather than exploring per-pixel segmentation techniques and datasets. 

In \citep{basu2015}, we presented a classification framework that feeds handcrafted features extracted from an image  to a DBN for classifying high resolution satellite imagery. The framework in \citep{basu2015} classifies satellite imagery without considering the spatial features or correlation information from the image. In this paper,  we present an improved architecture that enhances  a modern CNN  with handcrafted features for classification of high resolution satellite imagery. The framework presented in this paper fuses handcrafted features extracted from an image with spatial (deep) features acquired from the bottleneck layer of a  CNN to obtain improved classification accuracy on the SAT-4 and SAT-6 datasets compared to \citep{basu2015}.  

\begin{figure*}
  \centering
    \includegraphics[width=0.99\textwidth]{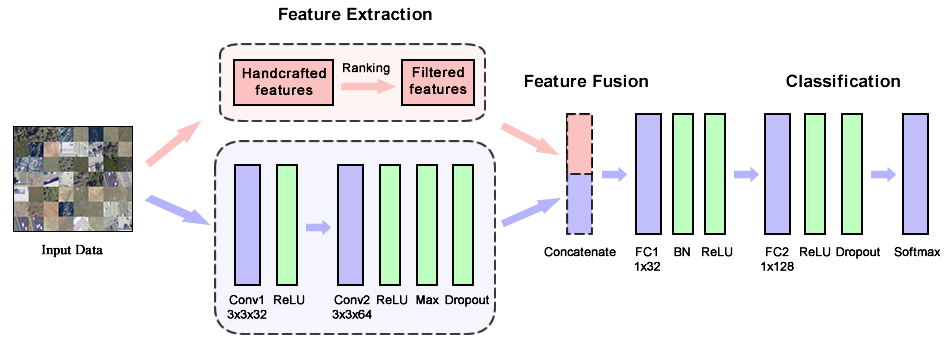}
  \caption{Architecture of the DeepSat V2 classification framework.} \label{arch}
\end{figure*}

\section{Architectural Overview}
We propose an end-to-end framework   that augments a modern CNN architecture with handcrafted features (texture features)  to improve distribution separability for classification of satellite imagery. While the DBN-based architecture in \citep{basu2015} used   higher-order texture features  that are important  for discriminative representations  for various landcover classes, it did not capture spatial contextual information. We extend \citep{basu2015} by providing a new architecture that uses a  CNN as a  baseline model  for extracting spatial contextual information and  then augmenting it with  the representations extracted from handcrafted feature spaces to enhance the discriminative power.

The complete architecture  is depicted in Figure \ref{arch}. It consists of  two convolutional layers with 32 and 64 feature maps with a kernel of 3$\times$3 for both, each accompanied with a Rectified Linear Unit (ReLU) layer. A max-pooling layer follows that with a kernel of 2$\times$2. A Dropout layer  is  added after the max pooling layer with dropout rate of  0.25. This is followed by a feature fusion layer where  the handcrafted features are concatenated with the  CNN bottleneck representations. Then the fused features are  input into a  fully connected dense  layer  containing 32  neurons to which batch normalization is added. Following this is a ReLU layer, after which is a fully connected dense layer with  128 neurons. After this layer comes a ReLU layer,   succeeding which is a   dropout layer with  rate   0.2.  The final layer is  a Softmax layer based on cross-entropy loss function. The Adadelta optimizer \citep{zeiler2012adadelta} have been adopted in the framework. 

\subsection{Feature Extraction}
The feature extraction phase computes 150 features from the input imagery.  The key features that we use for classification are mean, standard deviation, variance, 2nd moment, direct cosine transforms, correlation, co-variance, autocorrelation, energy, entropy, homogeneity, contrast, maximum probability and sum of variance of the hue, saturation, intensity, and near infrared (NIR) channels as well as those of the color co-occurrence matrices. These features were shown to be useful descriptors for classification of satellite imagery in previous research \citep{haralick1973}. Since two of the classes in SAT-4 and SAT-6 are trees and grasslands, we incorporate features that are useful determinants for segregation of vegetated areas from non-vegetated ones. The red band already provides a useful feature for discrimination of vegetated and non-vegetated areas based on chlorophyll reflectance.  However, we also use derived features (vegetation indices derived from spectral band combinations) that are more representative of vegetation greenness -- this includes the Enhanced Vegetation Index (EVI) \citep{huete2002}, Normalized Difference Vegetation Index (NDVI) \citep{rouse1974} and Atmospherically Resistant Vegetation Index (ARVI) \citep{kaufman1992}. 



The performance of our learner depends to a large extent on the selected features. Some features contribute more than others towards optimal classification. The 150 features extracted are narrowed down to 22 using a feature-ranking algorithm based on Distribution Separability Criterion~\citep{Boureau10atheoretical}. Details of the feature ranking method along with the ranking for all the 22 features used in our framework are provided in Section \ref{Section:feature_ranking}.

\begin{table}[b]
\begin{center}
\footnotesize
\begin{tabular}{cccccc}
\hline
 \textbf{Rank} & \textbf{Feature} & \multicolumn{1}{c|}{$\boldsymbol{D_{\mathrm s}}$} & \textbf{Rank} & \textbf{Feature}        & $\boldsymbol{D_{\mathrm s}}$       \\ \hline
1&I CCM mean         & \multicolumn{1}{c|}{2.9403}              & 12   &   I std    &   0.7968    \\
2 & H CCM sosvh       & \multicolumn{1}{c|}{2.5413}             & 13   &   H std    &    0.7956  \\
3 & H CCM autoc     & \multicolumn{1}{c|}{2.1417}               & 14   &   H mean   &   0.7632  \\
4 & S CCM mean       & \multicolumn{1}{c|}{1.4099}              & 15   &   I mean   &    0.7541  \\
5 &  H CCM mean         & \multicolumn{1}{c|}{1.1237}           & 16   &   S mean   &     0.7268 \\
6 &   SR          & \multicolumn{1}{c|}{0.9424}                 & 17   &   I CCM covariance & 0.7228 \\
7 &   S CCM 2nd moment         & \multicolumn{1}{c|}{0.8354}    & 18   &   NIR mean &    0.6997\\
8 &   I CCM 2nd moment         & \multicolumn{1}{c|}{0.8354}    & 19   &   ARVI     &     0.6622 \\
9  &   I 2nd moment    & \multicolumn{1}{c|}{0.8345}            & 20   &   NDVI     &     0.6594\\
10   &  I variance         & \multicolumn{1}{c|}{0.8345}        & 21   &    DCT     &    0.5792 \\
11   &   NIR std         & \multicolumn{1}{c|}{0.7980}          & 22   &    EVI     &    0.3207 \\ 
\hline 
\end{tabular}
\end{center}
\caption{Ranking of features based on Distribution Separability Criterion for SAT-6. Here CCM refers to Color Cooccurrence Matrix \citep{boyda2017deploying}, DCT to Discrete Cosine Transform, sosvh to sum of sqaures for variance, autoc to autocorrelation, std to standard deviation.}
\label{table:Feature_ranking}
\end{table}


\subsection{A Statistical Perspective based on Distribution Separability Criterion}\label{statistical_perspective}
\begin{figure}[t]
\centering
\subfigure[]{\includegraphics[width=0.4\linewidth, keepaspectratio]{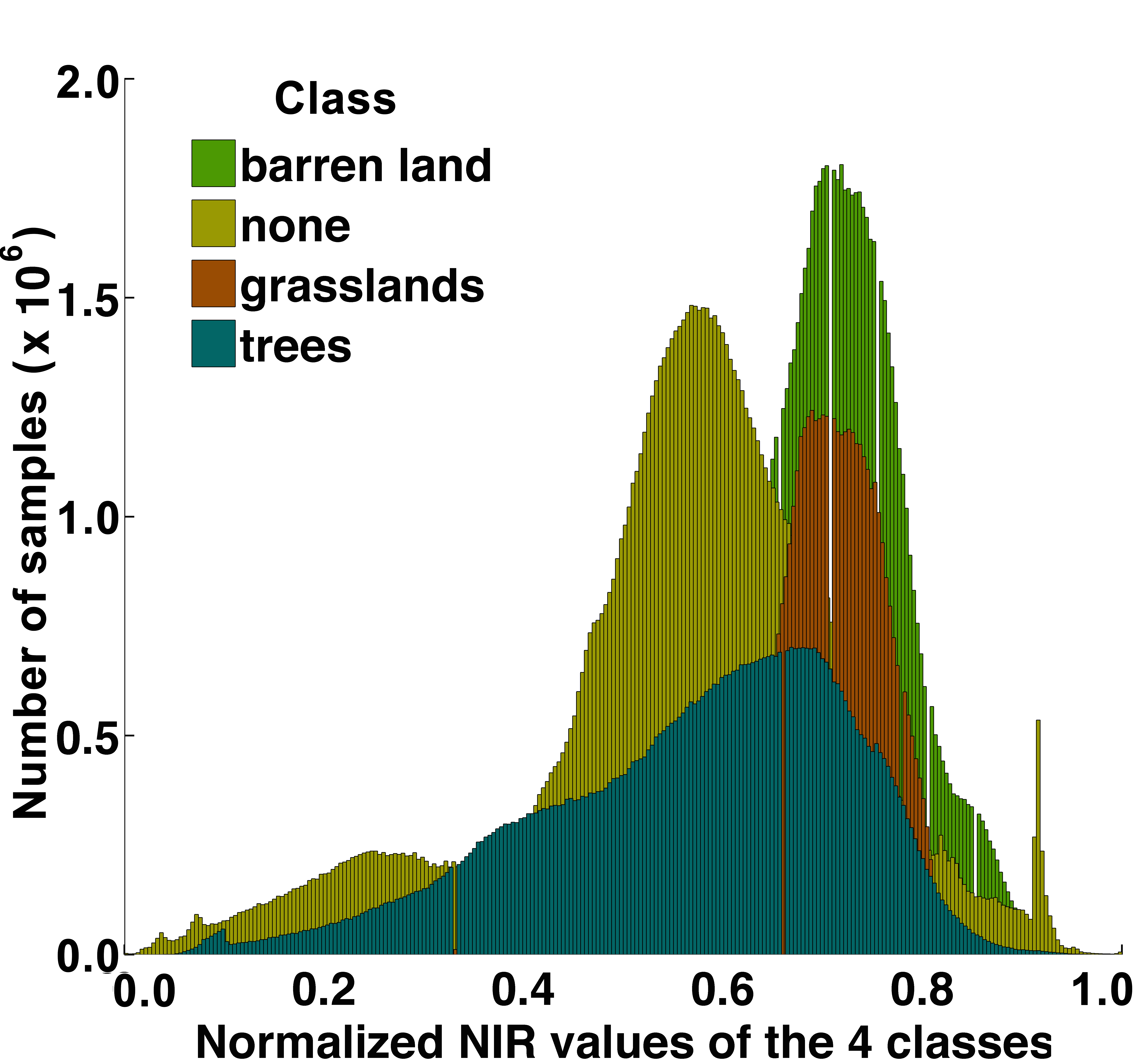}}
\hspace{10mm}
\subfigure[]{ \includegraphics[width=0.4\linewidth, keepaspectratio]{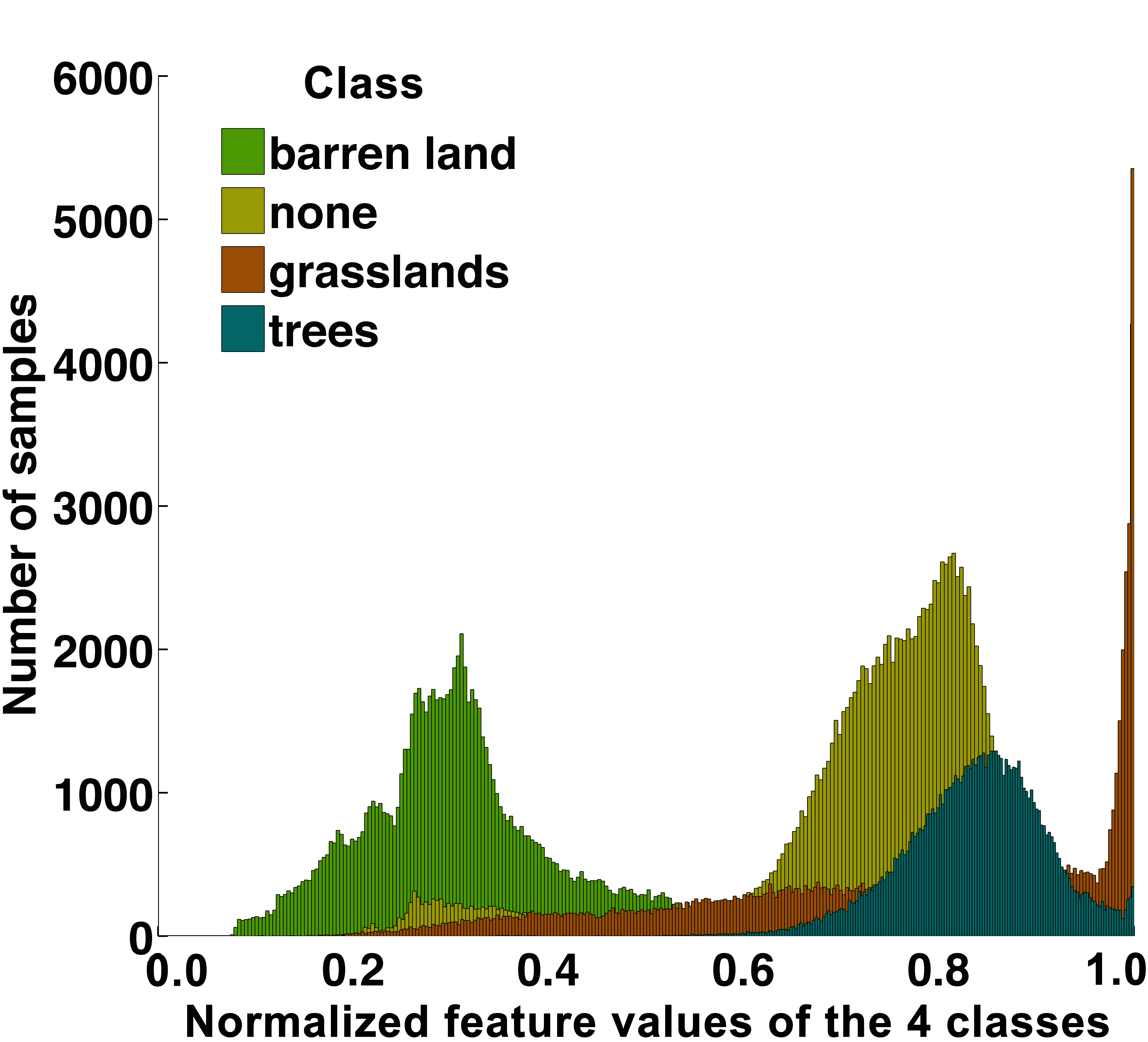}}
\caption{Distributions of the raw NIR values for traditional deep learning algorithms and a sample handcrafted DeepSat feature (Autocorrelation of Hue Color co-occurrence matrix \cite{boyda2017deploying}) for various classes in SAT-4 imagery.}
\label{fig:distributions}
\end{figure}

Improving classification accuracy can be viewed as maximizing the separability between the class-conditional distributions. We can view the problem of maximizing distribution separability \citep{Boureau10atheoretical} as maximizing the distance between distribution means and minimizing their standard deviations. Figure \ref{fig:distributions} shows the histograms that represent the class-conditional distributions of the NIR channel and a sample feature extracted in our framework. As illustrated in Table \ref{table:Distribution_mean_and_sd}, the features extracted in our framework have a higher distance between means and a lower standard deviation as compared to the original image distributions, thereby ensuring better class separability.

\begin{table}[b]
\begin{center}
\footnotesize
\begin{tabular}{ | c | c | c | c | }
    \hline
 \multirow{2}{*}{\textbf{Dataset}} & \multirow{2}{*}{\textbf{Type}} & \textbf{Distance between} & \textbf{Mean of Standard} \\ 
 &   & \textbf{Means} &   \textbf{Deviations}  \\ \hline
 \multirow{2}{*}{SAT-4} & Raw Images & 0.1994 & 0.1166 \\ 
  & Handcrafted DeepSat Features & 0.8454 & 0.0435 \\ \hline
 \multirow{2}{*}{SAT-6} & Raw Images & 0.3247 & 0.1273 \\
  & Handcrafted DeepSat Features & 0.9726 & 0.0491 \\ \hline
  \end{tabular}
  \end{center}
  \caption{Distance between Means and  Means of Standard Deviations for raw image values and DeepSat feature vectors for SAT-4 and SAT-6.}
  \label{table:Distribution_mean_and_sd}
\end{table}

\subsubsection{Feature Ranking}\label{Section:feature_ranking}
Following the analysis proposed in Section \ref{statistical_perspective} above, we can derive a metric for the Distribution Separability Criterion as follows:
$D_{\mathrm s} = \frac{\overbar{\lVert \delta_{mean} \rVert }}{\overbar{\delta_{\sigma}}}$
where $\overbar{\lVert \delta_{mean} \rVert }$ indicates the mean of distance between means and $\overbar{\delta_{\sigma}}$ indicates the mean of standard deviations of the class conditional distributions. Maximizing $D_{\mathrm s}$ over the feature space, a feature ranking can be obtained. Table \ref{table:Feature_ranking} shows the ranking of the various features used in our framework along with the values of the corresponding distance between means $\overbar{\lVert \delta_{mean} \rVert }$, standard deviation  $\overbar{\delta_{\sigma}}$, and Distribution Separability Criterion $D_{\mathrm s}$. A threshold of $D_{\mathrm s}=0.3$ was used to narrow down the $22$ features in Table \ref{table:Feature_ranking} from among $150$ features. 

\section{Experimental Results}
\subsection{Experimental Settings}

All of our experiments were conducted on  an Exxact workstation with one Intel Core i7-5930K CPU with 12 cores, four NVIDIA GeForce GTX TITAN X GPUs, and a 64 GB memory. The NVIDIA deep learning library of CuDNN of CUDA was used  for acceleration and our model was developed in Keras with  Tensorflow as backend.

\subsection{Performance Analysis}
We evaluated our architecture on the SAT-4 and SAT-6 datasets \citep{basu2015}. As stated above, among the publicly available benchmark datasets for high resolution satellite imagery in the remote sensing community \citep{ml}, only SAT-4 and SAT-6 provide enough labeled image patches (500,000 and 405,000 respectively) to evaluate a new architecture or approach without running into overtraining issues. The SAT-4 training set has 400,000 training samples of $28\times 28$ images each with $4$ channels \citep{basu2015} while the test set has 100,000 samples with the image size and channels remaining the same. The SAT-6 training set has 324,000 training samples of $28\times 28$ images each with $4$ channels \citep{basu2015} while the test set has 81,000 samples with the image size and channels remaining the same. 

\begin{figure}[t]
  \centering
    \includegraphics[width=0.90\textwidth]{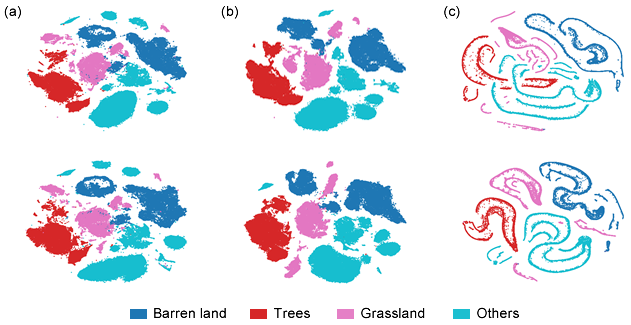}
  \caption{Visualization of learned representations and decision boundaries for SAT-4 dataset. Top row, regular CNN model which has no handcrafted features fused. Bottom, proposed framework which has handcrafted features fused. (a) Feature maps learned from the first dense layer. (b) Feature maps learned from the second dense layer. (c) Decision Boundaries.} \label{sne}
\end{figure}

To qualitatively understand the impact of augmentation  with handcrafted features, in  Figure \ref{sne}, we visualize the learned representations and the decision boundaries   for the SAT-4 dataset  using t-Distributed Stochastic Neighbor Embedding (t-SNE) \citep{maaten2008visualizing}, that embeds  representations in high dimensions into two dimensional space preserving  the  distances  based on local structure. To this end, t-SNE first generates a probability distribution over point pairs in high dimensional space using  a Gaussian distribution, ensuring that similar pairs have higher probability. It then generates the low dimensional mappings having the similar probability distributions  wherein similarity between points is estimated using  the student t-distribution. 
The bottom  row in  Figure \ref{sne} visualizes the map responses learned from the first fully connected dense layer, those learned from the second fully connected dense layer, and the decision boundaries, respectively,  for a CNN augmented with handcrafted features while the top row shows the same for the same CNN without the handcrafted features (and without the feature fusion layer).  It can be seen from Figure \ref{sne}   that  fusing  handcrafted features helped  improve discriminative  feature learning (see Figure \ref{sne}(B), bottom row,  where the others class is already  more compactly clustered than in the top)  providing robust separation of the decision boundaries  (see Figure \ref{sne}(C) where the bottom row shows clearer separation of the classes than the top where the classes trees, grassland, and others are not robustly separable and the intra-class distances are more). This is corroborated by the higher distances between means and the lower standard deviations for the handcrafted features as shown in Table \ref{table:Distribution_mean_and_sd}. 

\begin{figure}[t]
\centering
\subfigure[]{
    \includegraphics[width=0.46\columnwidth, keepaspectratio]{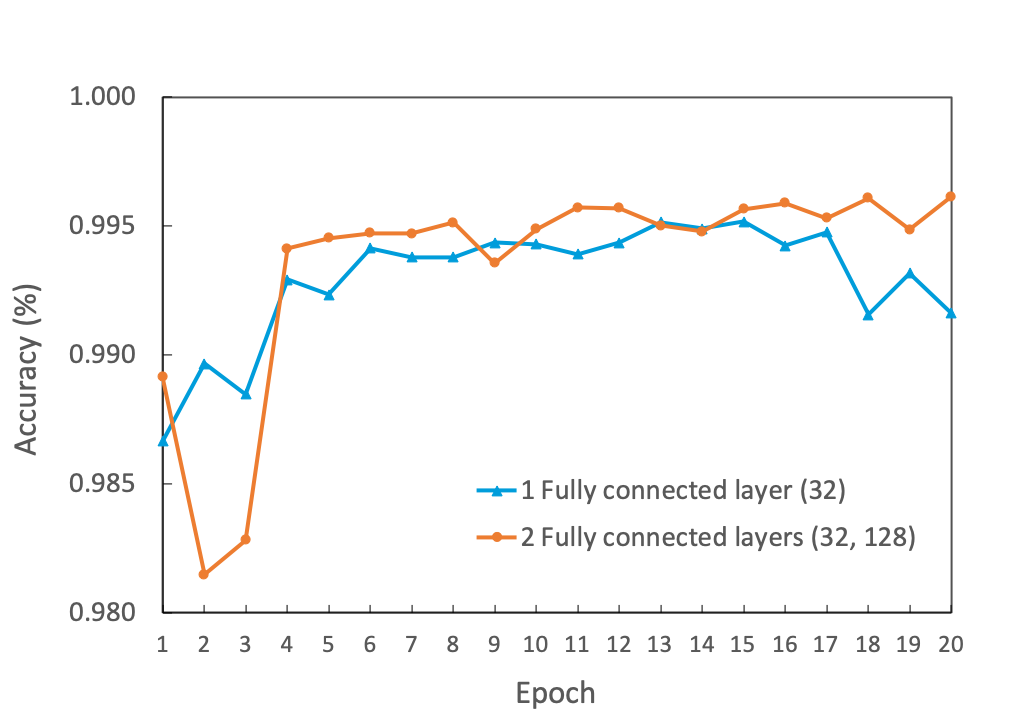}
    \label{fig:dense}
}
\hspace{10pt}
\subfigure[]{
    \includegraphics[width=0.46\columnwidth, keepaspectratio]{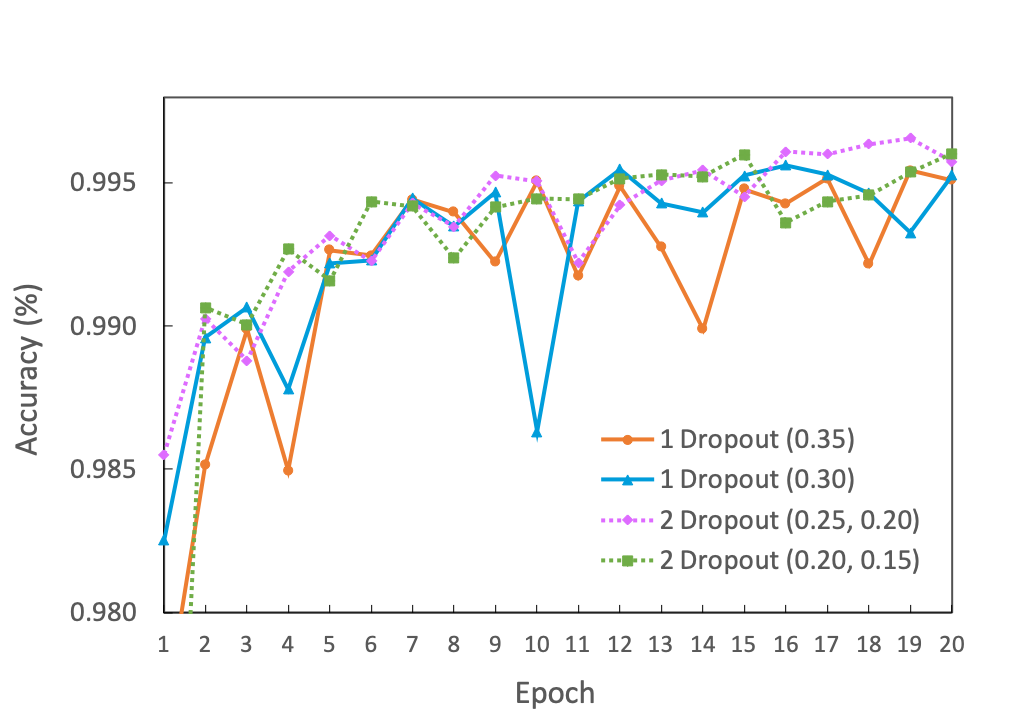}       
    \label{fig:dropout}
}

\caption[Optional caption for list of figures]{Impact on the classification performance of our framework on the datasets. (a) Using different number of fully connected layers with values. (b) Using different number of dropout layers with values.}
\label{fig:performance}
\end{figure}

We next study the impact of the two fully connected layers after the feature fusion layer as well as that of the dropout layers on classification (testing) accuracy in Figure \ref{fig:performance}. Both Figures \ref{fig:dense} and \ref{fig:dropout} show how classification  accuracy (testing) changes with  the  number of epochs. Figure \ref{fig:dense} shows that removing the second dense layer (with 128 neurons) reduces the  network  performance with respect to  accuracy of classification. Figure \ref{fig:dropout} shows that a dropout layer before the feature fusion layer with rate $0.25$ and one before the final layer with rate $0.2$ provides the best performance in terms of classification  accuracy (testing) (shown by pink line in Figure \ref{fig:dropout}).

\begin{table}[b]
\begin{center}
\footnotesize
\begin{tabular}{ |p{5.5cm}||>{\centering}m{2cm}|>{\centering}m{2cm}|}
 \hline
 Methods& SAT-4\\Accuracy (\%) & SAT-6\\Accuracy (\%)\tabularnewline
 \hline
 DBN~\citep{basu2015} & 81.78& 76.47 \tabularnewline
 SDAE~\citep{basu2015}   & 79.98& 78.43 \tabularnewline
 CNN~\citep{basu2015} & 86.83 & 79.10\tabularnewline
 DeepSat~\citep{basu2015}   & 97.95 & 93.92\tabularnewline
 Contrastive loss~\citep{simo2015discriminative}   & 98.74 & 98.55\tabularnewline
 MLP (\textit{Z}-score)~\citep{zhong2017satcnn}   & 94.76 & 97.46\tabularnewline
 DCNN~\citep{ma2016satellite} & 98.41 & 96.04\tabularnewline
 TradCNN (\textit{Z}-score)~\citep{zhong2017satcnn} & 98.43 & 98.34\tabularnewline
 D-DSML-CaffeNet~\citep{gong2018diversity} & 99.51 & 99.42\tabularnewline
 SatCNN (linear)~\citep{zhong2017satcnn} & 99.55 & 99.58\tabularnewline
 SatCNN (\textit{Z}-score)~\citep{zhong2017satcnn} & 99.69 & 99.61\tabularnewline
 Triplet networks~\citep{liu2018scene} & 99.76 & 99.71\tabularnewline
 \hline
 DeepSat V2 (The proposed method) & 99.90 & 99.84 \tabularnewline
 \hline
\end{tabular}
\end{center}
\caption {Comparison of classification accuracy (\%) of various methods on SAT-4 and SAT-6 datasets.} \label{comp}
\end{table}

\subsection{Comparison with State-of-the-Art Methods}
In this section, we compare the results obtained by using our  approach  with those obtained using   state-of-the-art methods  on  the SAT-4 and SAT-6 datasets. The comparison is  shown in Table \ref{comp}. The classification accuracy obtained using  our approach are 99.90\% on SAT-4 and 99.84\% on SAT-6.   
It can be seen from Table \ref{comp} that  our framework surpasses all the existing approaches in terms of accuracy of classification  \citep{basu2015,simo2015discriminative,zhong2017satcnn,ma2016satellite,gong2018diversity, liu2018scene}; in particular, it surpasses  the next best one \citep{liu2018scene}  that uses triplet networks by 0.14\% on SAT-4 and 0.13\% on SAT-6. We statistically evaluate the significance of the improvement provided by our framework over \citep{liu2018scene} using the McNemar's test (since the test datasets for  our framework and for \citep{liu2018scene} were same for both SAT-4 and SAT-6). For the SAT-4 dataset, using McNemar's test, we obtain  the value of the test statistic $\chi^2 = 138.01$ with degree of freedom $1$ and a two-tailed \textit{p}-value less than $2.2 \times 10^{-16}$ indicating that the improvement in the accuracy of classification induced by our framework is statistically significant.  For the SAT-6 dataset, using McNemar's test, we obtain  the value of the test statistic $\chi^2 = 103.01$ with degree of freedom $1$ and a two-tailed \textit{p}-value less than $2.2 \times 10^{-16}$ indicating that the improvement in the accuracy of classification induced by our framework is statistically significant. Our approach achieves better performance than that achieved by  complex triplet networks \citep{liu2018scene}  by augmenting  a smaller  CNN,  comprising  only  of  two convolutional layers together with  two fully connected  layers apart from  ReLU, Max-pooling, Dropout, and Softmax, with handcrafted features. The advantages of our framework are simplicity and fast training (with average training time being around 1200 seconds for both datasets as opposed to ${\sim}2400$ seconds for \citep{zhong2017satcnn}). 

\section{Conclusions}
We present an end-to-end framework based on an improved architecture that  augments a  CNN architecture with handcrafted features, for high resolution satellite imagery classification. We showed that augmenting a CNN with  handcrafted features  enhances its discriminative power  for satellite imagery even compared to larger unaugmented CNN architectures \citep{zhong2017satcnn} (see Table \ref{comp}). Our framework outperforms all the existing approaches \citep{basu2015,simo2015discriminative,zhong2017satcnn,ma2016satellite,gong2018diversity, liu2018scene} in terms of classification accuracy for the SAT-4 and SAT-6 datasets. A statistical analysis based on Distribution Separability Criterion  substantiates the robustness of our approach in learning better representations for satellite imagery.

\bibliographystyle{tfcad}
\bibliography{bibfile_Qun,ref}

\end{document}